\documentclass{article}
\usepackage{iclr2027_conference,times}

\usepackage{amsmath,amsfonts,bm}

\def\eqref#1{equation~\ref{#1}}
\def\1{\bm{1}}

\DeclareMathAlphabet{\mathsfit}{\encodingdefault}{\sfdefault}{m}{sl}
\SetMathAlphabet{\mathsfit}{bold}{\encodingdefault}{\sfdefault}{bx}{n}

\usepackage{hyperref}
\usepackage{url}
\usepackage{xspace}
\usepackage{amsmath,amssymb}
\usepackage{amsthm}
\usepackage{booktabs}
\usepackage{multirow}
\usepackage{graphicx}
\usepackage{xcolor}

\renewcommand{\eqref}[1]{\textup{(\ref{#1})}}
\newcommand{\method}{Polar-ViT\xspace}
\newtheorem{proposition}{Proposition}

\title{Rethinking the Fully Hyperbolic Vision Transformer in Polar Coordinates}

\author{Ahmad Bdeir \& Niels Landwehr\\
Data Science Group\\
University of Hildesheim\\
Hildesheim, Germany \\
\texttt{\{bdeira,landwehr\}@uni-hildesheim.de} \\
}

\iclrfinalcopy
\begin{document}

\maketitle

\begin{abstract}

Hyperbolic space can embed intrinsic hierarchies in data with low distortion due to the exponential growth of volume with distance from the origin. However, current Lorentz transformer blocks are formulated in ambient coordinates, where numerical errors increase at large radii due to instability in the Lorentzian inner product, leading many models to limit the radius to avoid this issue. This prevents us from utilizing the regions of hyperbolic space that motivate the geometry. To address this, we revisit the components of the transformer block in polar coordinates, where hyperbolic operations such as distance calculation and attention centroids can be computed without the numerical cancellation in their ambient-coordinate formulations. Specifically, we propose a polar fully connected layer that separately maps an embedding's direction and radius, allowing the radius of a feature to be learned rather than determined by the norm of a linear map. We additionally introduce horospherical shifts as relative positional encodings whose query-key distances grow logarithmically with the token gap. Finally, we reformulate the residual connection as a Lorentz boost, which is an isometry at any radius. Combining these components, we develop a fully hyperbolic transformer that substantially improves performance over Euclidean and hyperbolic baselines on standard vision tasks. We further evaluate our model on ImageNet and demonstrate its ability to generalize to other datasets using pre-trained weights, similar to Euclidean counterparts.
\end{abstract}

\section{Introduction}
The use of hyperbolic geometry has become more prevalent in both the deep learning architectures and the tasks they are applied to.
 Hyperbolic manifolds are able to embed hierarchical and relational structure with low distortion because the volume
of the space grows exponentially with the distance from the origin. This makes it similar to the growth of trees with
depth~\citep{krioukov2010hyperbolic, sarkar2011low}.
However, the mechanism for volume growth also implies that the representational capacity of hyperbolic space greatly benefits from the ability to embed points at a large distance from the origin. But \cite{mishne2023numerical} show that the different models of hyperbolic space, mainly the Lorentz model and the Poincar\'e model, can only represent points accurately within certain radii that are fixed by machine precision. For the Lorentz model, this is approximately $19$ in double precision and extending their
argument to single precision gives us a radius of approximately $8$. Beyond this, the Lorentz inner product,
distance and weighted centroid operations lose their accuracy and are greatly affected by numerical instability~\citep{mishne2023numerical, bdeir2025robust}.

To avoid this, existing models confine their embeddings to a small radius through strict normalization
in the fully connected layer~\citep{chen2022fully}, norm clipping after
projection~\citep{yang2024hypformer, bdeir2024hcnn}, and frequent rescaling~\citep{bdeir2025robust}. This mitigates the instability but also prevents us from utilizing more of the volume that motivates the geometry. Additionally, this failure mode is
silent in the Lorentz model \citep{mishne2023numerical} since the operations do not directly become undefined. Instead, they return incorrect points which lead to training degradation or divergence without any component being identifiable as the cause~\citep{alyoussef2026hexformer}. \cite{alyoussef2026hexformer} identify this issue in the attention aggregation of hyperbolic transformers in particular and demonstrate the performance improvement of avoiding the operation entirely and replacing it with tangent-space aggregation~\citep{alyoussef2026hexformer, feinashley2024hvt}.

Additionally, most of the hyperbolic Transformers rely on Euclidean components for positional encodings, residual connections, and layer normalization. These operations are typically performed on the space components of a hyperbolic embedding and then the time component is recomputed from the manifold constraint. Despite being effective, this approach introduces another source of numerical instability. The time component grows exponentially with the radius which makes it subject to cancellation errors when being recomputed from the space values.

Our work introduces \method, a hyperbolic transformer that relies on polar form operations, representing each embedding by its geodesic radius and unit direction. When using polar coordinates, we show that the terms that cancel in ambient Lorentz coordinates are never formed, so the weighted centroid and the distance remain accurate at radii past the ambient representation limit. This lessens the need for normalization, clipping and rescaling components and allows us to use the space more effectively. Building on this representation, we additionally propose a polar fully connected layer that uses this representation to map direction and radius as separate channels. In terms of the remaining Euclidean components, we propose intrinsic hyperbolic residual and layer norm alternatives. We also present an argument for augmenting Euclidean positional encodings with a relative encoding derived from horospherical shifts, the parabolic subgroup of the Lorentz group. This approach shows increased model performance while encoding positions without recomputing any time coordinate. We evaluate \method at Tiny, Small and Base scales and release ImageNet-1k~\citep{deng2009imagenet} pre-trained weights to demonstrate their ability to transfer to downstream fine-tuning tasks similar to Euclidean Transformers~\citep{dosovitskiy2021vit}. The main contributions are then as follows:
\begin{itemize}
  \item Precision analysis: We show that the ambient weighted centroid fails beyond a radius of $8.3$ in single precision and derive a polar form that stays accurate up to a radius of about $17.3$.
  \item Polar hyperbolic Transformer (\method): A Transformer based on the
  polar form of the Lorentz model that supports embeddings at higher radii, along with a polar fully connected layer.
  \item Horospherical positional encodings: A relative positional encoding for hyperbolic
  attention built from horospherical shifts, which improves over Euclidean encodings alone.
  \item Intrinsic alternatives to residual connections and Layer Normalization.
  \item Pre-trained ImageNet-1k weights: Tiny and Small ViT checkpoints will be released,
  and we show that they transfer effectively to fine-tuning tasks, lowering the cost of adopting
  hyperbolic models.
\end{itemize}

\section{Related Work}

\textbf{Hyperbolic Neural Networks} Early work on hyperbolic embeddings and deep learning used the Poincar\'e
ball model of hyperbolic space~\citep{nickel2017poincare, ganea2018hyperbolic, shimizu2021hnnpp}. In parallel, the Lorentz model~\citep{nickel2018learning} followed a similar path with hybrid approaches until \cite{chen2022fully} proposed their fully hyperbolic network that defined a Lorentz linear layer and other components needed for hyperbolic encoders. The Lorentz model was then used to develop hyperbolic convolutional blocks, residual connections and normalization layers and applied to vision tasks~\citep{bdeir2024hcnn, he2025lorentzian, shi2026intrinsic,
vanspengler2023poincare, mishra2026hyperbolic} and natural language tasks~\citep{yang2024hyperbolic, he2025helm}. However, these models still rely on storing the
embeddings in ambient coordinates, which limits the radius that keeps Lorentzian operations well-defined~\citep{mishne2023numerical}.

\textbf{Hyperbolic Transformers} Hyperbolic attention networks~\citep{gulcehre2019hyperbolic}
were the first to calculate attention scores using distances on the hyperboloid. \cite{chen2022fully} then kept the whole attention layer in the
Lorentz model. They replaced the distance scores with squared distances and replaced the Einstein midpoint with the weighted Lorentzian centroid~\citep{law2019LorentzianDL}. As for vision tasks, HVT~\citep{feinashley2024hvt} replaced the attention and feed-forward layers with M\"obius operations in the Poincar\'e ball and trained on ImageNet-1k. For the Lorentz model, LViT from HyperCore~\citep{he2025hypercore} builds a fully hyperbolic Lorentz ViT using existing Lorentzian components. They are also the first to train a fully hyperbolic ViT on ImageNet-1k, but they use non-standard fine-tuning tests (using HypLoRA~\citep{yang2024hyperbolic}) to transfer to CIFAR and Tiny-ImageNet. HexFormer~\citep{alyoussef2026hexformer} argues for tangent space aggregation in the attention layer due to the numerical precision issues with the Lorentzian centroid. In comparison, our method keeps the weighted centroid by formulating it in a more exact setting and introduces improved hyperbolic components to increase performance. Additionally, none of the previous ViTs release ImageNet-1k weights,
and a recent survey~\citep{he2025survey} lists pre-trained hyperbolic vision models as still
missing and stresses their importance for easier adoption.

\textbf{Numerical Stability of Hyperbolic Representations} The number of bits needed to store a
hyperbolic point grows with its radius~\citep{sala2018representation}. Prior fixes fall into three groups. The
first group attempts to increase the precision through tiling the space~\citep{yu2019numerically} or through multi-component floats~\citep{yu2021representing, yu2022mctensor} which bound the error at any radius. The second group works on bounding the radius through clipping features before the exponential map~\citep{guo2022clipped}, fixing a radius threshold~\citep{skopek2020mixed}, normalizing inside the linear layer~\citep{chen2022fully}, or scaling the embeddings according to the largest representable distance for the current curvature value and precision~\citep{bdeir2025robust}. The third group changes the parametrization entirely. This is done through storing points in the
tangent space at the origin~\citep{mathieu2019continuous}. Closest to our
parametrization, several of the graph embedding methods that \citet{perezcasulo2026unified} unify in one framework
fit node positions directly in polar coordinates, so
the polar form has to cover only distances. \method extends the use of polar coordinates to the main
operations of a transformer, including the linear layer, attention and more importantly, the centroid.

\textbf{Positional Encodings} Relative positional encodings make the attention score depend on
the index gap~\citep{shaw2018self, press2022train}. Rotary position embedding~\citep{su2024roformer} rotates each
query and key by an angle that grows with its index, so the inner product of the pair depends
only on their relative position. Two recent works replace the rotation of rotary encoding with a Lorentz transformation. HELM~\citep{he2025helm} rotates the
space coordinates and recomputes the time coordinate, and HoPE~\citep{dai2025hope} applies a Lorentz boost to the Euclidean queries and keys of a standard
language model. Our approach uses the third, horospherical shifts, which slide points along horospheres
around a fixed ideal point. Horospheres have previously been used for prototypes~\citep{ghadimiatigh2021hyperbolic},
decision boundaries~\citep{fan2023horospherical}, dimensionality reduction~\citep{chami2021horopca}
and the linear layer~\citep{chen2026hyperbolic}, but not for positional encoding.

\section{The Lorentz Model}
\label{sec:manifold}

\paragraph{Ambient coordinates.}
The $n$-dimensional Lorentz model with curvature $-1/k$ is the upper sheet of the hyperboloid defined as
$\mathbb{L}^n_k = \{x \in \mathbb{R}^{n+1} : \langle x, x \rangle_{\mathcal{L}} = -k,\ x_0 > 0\}$.
Here, the hyperbolic point is $x = (x_0, \mathbf{x})$ with time coordinate $x_0$ and space coordinates
$\mathbf{x} \in \mathbb{R}^n$, and $\langle x, y \rangle_{\mathcal{L}} = -x_0 y_0 + \mathbf{x} \cdot \mathbf{y}$
is the Lorentzian inner product. Based on the definitions, the time coordinate can be computed from the space coordinates as
$x_0 = \sqrt{k + \|\mathbf{x}\|^2}$. Applying this to a space vector of all 0s gives the origin $o = (\sqrt{k}, 0, \dots, 0)$. Finally, the Lorentzian distance $d_{\mathcal{L}}(x, y)$ is the length of the geodesic connecting two points on the hyperboloid $x$ and $y$,
\begin{equation}
  d_{\mathcal{L}}(x, y) = \sqrt{k}\, \operatorname{arcosh}\!\big(-\langle x, y \rangle_{\mathcal{L}} / k\big).
  \label{eq:dist}
\end{equation}

\paragraph{Polar form.}
Instead of the standard time and space coordinate representation, we can determine a point by its polar form $(r, \mathbf{u})$, where $r = d_{\mathcal{L}}(o, x)$ is the distance from the origin and $\mathbf{u} = \mathbf{x}/\|\mathbf{x}\|$ is the unit direction of its space coordinates. Applying Eq.~\eqref{eq:dist} to the origin gives $r = \sqrt{k}\operatorname{arcosh}(x_0/\sqrt{k})$, so $x_0 = \sqrt{k}\cosh(r/\sqrt{k})$, and the constraint $x_0^2 - \|\mathbf{x}\|^2 = k$ then gives $\|\mathbf{x}\| = \sqrt{k}\sinh(r/\sqrt{k})$. We can therefore convert back to ambient coordinates by
\begin{equation}
  x = \Big(\sqrt{k} \cosh\tfrac{r}{\sqrt{k}},\; \sqrt{k} \sinh\tfrac{r}{\sqrt{k}}\, \mathbf{u}\Big),
  \label{eq:polar}
\end{equation}
and from ambient to polar by $r = \sqrt{k}\, \operatorname{arsinh}(\|\mathbf{x}\| / \sqrt{k})$ and $\mathbf{u} = \mathbf{x}/\|\mathbf{x}\|$.
Throughout, we write $a = r/\sqrt{k}$ for the rescaled radius, so that $x = (\sqrt{k}\cosh a, \sqrt{k}\sinh a\, \mathbf{u})$.
Both ambient coordinates grow as $e^{a}/2$, so their rounding error grows as $\varepsilon e^{a}$
for a unit roundoff $\varepsilon$. The radius and the direction, in contrast, are stored to
relative precision $\varepsilon$ at every radius. \method stores every embedding in polar form
$(r, \mathbf{u})$ and evaluates the sensitive operations in it.

\subsection{Precision Loss in Ambient Coordinates}
\label{sec:sensitive}
Most Lorentz operations rely on the Lorentzian inner product, either directly (distance,
logarithmic map, parallel transport) or through a derived form (centroid).
\citet{mishne2023numerical} bound the radius at which the coordinates of a single point can still be stored accurately. We extend
this analysis to the distance and the centroid calculation, the two operations Lorentzian attention heavily relies on. In the following, we denote $\varepsilon$ as the unit roundoff: $2^{-24}$ in single precision, $2^{-53}$ in
double precision and $2^{-8}$ in bfloat16.

\paragraph{Cancellation in the inner product.}
Consider two points $x$ and $y$ on the hyperboloid with rescaled radii $a$ and $b$, and whose directions form an angle $\theta$.
Then Eq.~\eqref{eq:polar} gives
\begin{equation}
  \frac{x_0 y_0}{k} = \cosh a \cosh b, \qquad
  \frac{\mathbf{x} \cdot \mathbf{y}}{k} = \sinh a \sinh b \cos\theta, \qquad
  -\frac{\langle x, y \rangle_{\mathcal{L}}}{k} = \cosh a \cosh b - \sinh a \sinh b \cos\theta .
  \label{eq:minkowski-polar}
\end{equation}
The first two terms both grow as $e^{a+b}/4$, while their difference,
$\cosh(d_{\mathcal{L}}(x, y)/\sqrt{k})$, stays close to one for nearby points. Each product is
rounded to the relative precision $\varepsilon$, so the computed inner product carries an absolute
error of around $\varepsilon e^{a+b}/4$. This error is independent of $d_{\mathcal{L}}(x, y)$ and depends mainly on the radii of the points.
\citet{mishne2023numerical} give the simplest instance of this effect. For a point at radius $\approx 19.1$ and $k = 1$,
$x_0 = 10^8$, $x_0^2 = 10^{16}$, and in double precision $x_0^2 - 1$ rounds to
$x_0^2$, so $\langle x, x \rangle_{\mathcal{L}}$ returns $0$ instead of $-1$.

\paragraph{Distance.}
By Eq.~\eqref{eq:dist}, we can find the normalized inner product with respect to the Lorentzian distance as $-\langle x, y \rangle_{\mathcal{L}}/k
= \cosh(d_{\mathcal{L}}(x, y)/\sqrt{k})$, which is always at least one. We define the term
$c = \cosh(d_{\mathcal{L}}(x, y)/\sqrt{k}) - 1$ as the difference from this minimum amount. Since
$\cosh t - 1 \approx t^2/2$ for small $t$, nearby points have $c \approx d_{\mathcal{L}}^2(x, y)/2k$,
so all information about the distance is based on this $c$. The absolute rounding error of the inner
product is added directly to $c$. For two points at the same radius $a$ this resolves to $\varepsilon e^{2a}/4$. As such we can see that once the error is larger than $c$ itself, the computed distance is meaningless.

For example, take $k = 1$ and two points at radius $a = 10$ that are $d_{\mathcal{L}}(x, y) = 1$
apart. Their ambient coordinates are $x_0 = y_0 = \cosh 10$ and
$\|\mathbf{x}\| = \|\mathbf{y}\| = \sinh 10$. The ambient
operations evaluate $-\langle x, y \rangle_{\mathcal{L}} = x_0 y_0 - \mathbf{x} \cdot \mathbf{y}$, where
$x_0 y_0 \approx 1.2129 \times 10^{8}$ and $\mathbf{x} \cdot \mathbf{y} \approx 1.2129 \times 10^{8}$. Here the true difference should be $\cosh 1 = 1.54$. In single precision each of these two products is
rounded to about seven significant digits, so each carries an absolute error of order
$2^{-24} \times 1.2 \times 10^{8} \approx 7$. If we perform the subtraction in single precision we get around $24$ and this is justified because the error is the same size as the error on the inputs. This value is several times larger than $1.54$ and the distance between $x$ and $y$ becomes $\operatorname{arcosh} 24 = 3.87$ instead of $1$. At $a = 12$ the two products are
$e^{4}$ times larger and their errors reach about $400$. The subtraction returns $0$, and
$\operatorname{arcosh} 0$ is NaN.

\begin{proposition}[Ambient failure radius]
\label{prop:ambient}
In ambient coordinates with unit roundoff $\varepsilon$, the rounding error on the computed
distance between two points at radius $r$ exceeds the distance itself, for distances of order
$\sqrt{k}$, once $r \geq r_\varepsilon = \tfrac{\sqrt{k}}{2} \ln(1/\varepsilon)$.
In the case of $k = 1$, this is $r_\varepsilon = 8.3$ in single precision, $18.4$ in double precision and $2.8$ in
bfloat16.
\end{proposition}

The double-precision value agrees with the bound of \citet{mishne2023numerical}, and the
single-precision value agrees with the radius at which we measure a $100\%$ error in the ambient distance
(Appendix~\ref{sec:exp-precision}). The logarithmic map, parallel transport and the tangent inner
product of Riemannian optimizers use the same inner product and fail at the same radius.

\paragraph{Centroid.}
The weighted centroid~\citep{law2019LorentzianDL} rescales the weighted sum of its inputs onto
the hyperboloid,
\begin{equation}
  \mu(x, w) = \sqrt{k}\, \frac{\sum_i w_i x_i}{\big\|\sum_i w_i x_i\big\|_{\mathcal{L}}},
  \qquad
  \|v\|_{\mathcal{L}} = \sqrt{|\langle v, v \rangle_{\mathcal{L}}|}.
  \label{eq:centroid}
\end{equation}
This operation is important because it aggregates the values in Lorentz attention~\citep{chen2022fully, yang2024hypformer}, and it is used to calculate the mean in Lorentz batch normalization~\citep{bdeir2024hcnn}, the Lorentzian
residual~\citep{he2025lorentzian} and the manifold weight decay of Riemannian
AdamW~\citep{bdeir2025robust}. However, the denominator is based on the ambient inner product and is then also subject to silent or absolute failure.

Let $T = \sum_i w_i \cosh a_i$ and
$\mathbf{V} = \sum_i w_i \sinh a_i\, \mathbf{u}_i$ be the time and space parts of the weighted
sum, divided by $\sqrt{k}$. The square of the denominator is then $T^2 - \|\mathbf{V}\|^2$.
Ambient operation computes $T^2$ and $\|\mathbf{V}\|^2$ separately and since the size of each is about
$e^{2a}/4$, their difference carries an error of about $\varepsilon e^{2a}/4$. The true
difference is much smaller since it is actually of order $e^{a}$ times a quantity that shrinks as the points
move outward and as their directions align. Once the rounding error reaches the true value, the
computed normalizer is noise.

For example, manifold weight decay pulls a parameter $x$ towards the origin $o$ by taking the
centroid of $\{x, o\}$ with weights $(1 - w, w)$, where $w$ is small~\citep{bdeir2025robust}. Take $k = 1$, $w = 10^{-5}$
and $x$ at radius $a = 8$. The step should move $x$ towards the origin to the radius $7.985$. However, the two
squared terms are $T^2 \approx \|\mathbf{V}\|^2 \approx 2.2 \times 10^{6}$ and single precision returns the difference as
$0.75$, which puts the output at radius $8.14$. This means that the decay step pushed the parameter in the opposite direction.
At $a = 9$ both squares round to an identical value, so the subtraction returns $0$ and the division returns NaN.

\subsection{Polar Operations and Their Precision}
\label{sec:polar-ops}
Every failure mode above shares the same pattern: two large ambient quantities are computed separately and
then subtracted. In polar form, we expand each operation before evaluating it, so that the large
parts cancel out, and the operation becomes a sum of non-negative terms.

\paragraph{Distance formulation.}
Subtracting one from Eq.~\eqref{eq:minkowski-polar} and using
$1 - \cos\theta = \tfrac12\|\mathbf{u} - \mathbf{v}\|^2$ gives the hyperbolic law of haversines,
\begin{equation}
  c = \cosh\!\big(d_{\mathcal{L}}(x, y)/\sqrt{k}\big) - 1
  = 2 \sinh^2\!\big(\tfrac{a - b}{2}\big) + \sinh a \sinh b\, \tfrac12\|\mathbf{u} - \mathbf{v}\|^2,
  \qquad
  d_{\mathcal{L}}(x, y) = \sqrt{k}\, \operatorname{arsinh}\sqrt{c\,(c + 2)}.
  \label{eq:polar-dist}
\end{equation}
The two terms depend only on the radial and the angular gap and do not involve subtraction. This identity is classical, and we use it in place of the ambient inner product. Appendix~\ref{app:impl}
shows that a closed-form gradient ensuring the term is not evaluated as $1 - \mathbf{u} \cdot \mathbf{v}$ does not incur any extra computational cost or memory footprint over ambient.

\paragraph{Centroid formulation.}
The centroid has no such classical form, so we derive one. We factorize the squared denominator as
$T^2 - \|\mathbf{V}\|^2 = (T - \|\mathbf{V}\|)(T + \|\mathbf{V}\|)$. The second factor is a sum of
non-negative terms which is not problematic. We decompose the first factor using $\hat{\mathbf{u}} = \mathbf{V}/\|\mathbf{V}\|$ as:
\begin{equation}
  T - \|\mathbf{V}\| = \underbrace{\textstyle\sum_i w_i e^{-a_i}}_{E}
                     + \underbrace{\textstyle\sum_i w_i \sinh a_i\, \tfrac{1}{2}\|\mathbf{u}_i - \hat{\mathbf{u}}\|^2}_{S}.
  \label{eq:split}
\end{equation}
Here, $E$ is a radial term that shrinks as the points move outward, and $S$ is an angular term that
measures the spread of the directions of the inputs and vanishes when they coincide. The true
squared denominator is then $\approx (E + S)\, e^{a}$, and comparing it with the ambient error
of $\varepsilon e^{2a}/4$ shows that the ambient centroid is only correct given
$E + S \gtrsim \varepsilon\, e^{a}$.
Both factors, $T + \|\mathbf{V}\|$ and $E + S$, are sums of non-negative terms which are now more stable, and the polar
centroid computes them directly from the radii and directions. Its output has a direction of $\hat{\mathbf{u}}$ and a rescaled radius of
\begin{equation}
  \sinh a_\mu = \frac{\|\mathbf{V}\|}{\sqrt{(E + S)(T + \|\mathbf{V}\|)}} .
  \label{eq:polar-centroid}
\end{equation}

\paragraph{Tangent vectors and precision bound.}
The tangent-space inner product used by Riemannian optimizers can also be reduced to a plain sum in the same way, so the
optimizer also works in polar form. We show in Proposition~\ref{prop:polar} in
Appendix~\ref{app:polar-ops} that every polar operation is accurate to the limit of the accuracy of the stored points. This gives a radius of around $r \approx 17.3$ in single precision (against $8.3$ for ambient) and $6.2$ in bfloat16 (against $2.8$).

\section{Methodology}
\label{sec:method}
\method is a hyperbolic Vision Transformer that stores embeddings in polar form and evaluates
the sensitive operations with the more stable formulations proposed in Section~\ref{sec:polar-ops}. Image patches are first mapped onto
the manifold and embedded by a polar fully connected layer, and a learned absolute position table
is added to the space coordinates~\citep{alyoussef2026hexformer}. Each of the $L$ Transformer
blocks applies gyro layer normalization, a multi-head attention whose queries and keys carry the
horospherical positional encoding, and a two-layer MLP. Boost
residuals are used to maintain the signal. Finally, the output tokens are pooled by their polar centroid, as in LViT~\citep{he2025hypercore},
and classified by a hyperbolic multinomial logistic regression~\citep{bdeir2024hcnn}.

\subsection{Polar Fully Connected Layer}
\label{sec:polar-fc}
The Lorentz fully connected layer of \citet{chen2022fully} uses standard matrix multiplication on the input with a learned weight. In its general form, it then recomputes the time coordinate from the manifold constraint, so a single map outputs both the direction and the radius. In the form they use in practice, a separate sigmoid of a linear function of the input sets the time coordinate, and the space coordinates are rescaled to match, so the radius is bounded by a fixed constant. In a layer expressivity test, \cite{vanderklis2026fast} show that the necessary number of optimization steps the general form needs to reach a target radius grows exponentially with the distance from the origin. This is due to the logarithmic scaling of the output hyperbolic norms with respect to the number of gradient descent steps.

Our Polar-FC avoids both the slow growth and the fixed bound by giving the radius its own unbounded parametrization. As in the Lorentz layer, we multiply the input with a learned map $\mathbf{W}$ and apply the optional activation $\phi$ to produce $\mathbf{h}$. The output direction is then the direction of $\mathbf{h}$. We then parametrize the output radius with a scalar $\ell$ by taking it to be the logarithm of $\sinh a'$. This gives us:
\begin{equation}
  \mathbf{h} = \phi\big(\mathbf{W} x / \sqrt{k}\big), \qquad
  \mathbf{u}' = \mathbf{h} / \|\mathbf{h}\|, \qquad
  a' = \operatorname{arsinh}\!\big(e^{\ell}\big), \quad
  \ell = \beta \ln\!\big(\|\mathbf{h}\|/\sqrt{k}\big) + \lambda a + \boldsymbol{\varphi}^\top \boldsymbol{\chi} + \varphi_0 .
  \label{eq:polar-fc}
\end{equation}

Here $\boldsymbol{\chi} = \tanh(a)\,\mathbf{u}$ is the Klein vector of the input, which equals the space coordinates divided by the time coordinate. We use this term because we want to include the input direction $\mathbf{u}$ in the radius computation but $\mathbf{u}$ has a gradient of $1/\|\mathbf{x}\|$ at the origin while $\boldsymbol{\chi}$ remains smooth there. The log radius $\ell$ then has three weighted contributions from the output norm controlled by $\beta$, the input radius controlled by $\lambda$,
and the input direction controlled by $\boldsymbol{\varphi}$, plus a bias $\varphi_0$. This last term is a
single linear unit that lets the layer push points inward or outward depending on where they
point, which the direction map $\mathbf{W}$ cannot do since it only sets $\mathbf{u}'$. At initialization, we set $\beta = 1$, $\lambda = \varphi_0 = 0$, and $\boldsymbol{\varphi} = 0$, which reduces Eq.~\eqref{eq:polar-fc} to the general form of the Lorentz fully connected layer. For an input on $\mathbb{L}^n_k$ and an output on $\mathbb{L}^{n'}_k$, the number of parameters becomes $n' \times (n + 1) + n + 3$ instead of $n' \times (n + 1)$, which is
a negligible overhead.

We repeat the expressivity test of \citet{vanderklis2026fast} in Appendix~\ref{app:exp-test} and show that the Polar-FC reaches the target radius in both Float64, the original testing precision, and Float32, which is the precision we use in our experiments. Additionally, we are able to achieve this without the need for clipping which increases the convergence speed dramatically. It should be noted that we had to use the polar distance formulation for the loss in both scenarios since the ambient distance would either fail or report false convergence. We note that FGG-LNN~\citep{vanderklis2026fast} was also able to reach the target radius with its hyperplane-based formulation, but it was less stable without gradient clipping. The instability showed only at radii past 13 so we still included it in the ablation study in Appendix~\ref{app:ablation}.

\subsection{Attention with Polar Centroid Aggregation}
\label{sec:attention}
We calculate the queries, keys and values using the proposed Polar-FC layers. For multi-head attention, we follow the Lorentz Transformer of
\citet{chen2022fully}. In their work, the space coordinates are split into $H$ slices, each with its own time
coordinate, so that every head works on $\mathbb{L}^{n/H}_k$. The head outputs are later merged using
Lorentz direct concatenation. We follow this process with its polar coordinate equivalent operations. The score $s_{ij}$ between query $q_i$ and key $\kappa_j$ is calculated as the negative squared distance, computed with Eq.~\eqref{eq:polar-dist}, and the values are aggregated using the polar centroid:
\begin{equation}
  s_{ij} = -\frac{\gamma}{\tau}\, d_{\mathcal{L}}^2(q_i, \kappa_j), \qquad
  \mathbf{p}_{i} = \operatorname{softmax}_j(s_{ij}), \qquad
  z_i = \mu_p\big(v, \mathbf{p}_{i}\big),
  \label{eq:attention}
\end{equation}
where $\gamma = (n/H)^{-1/2}$ is a scale parameter and $\tau$ is the learned temperature, and $\mu_p$ is the polar centroid function. We fuse the score, the softmax and
the centroid into a single kernel that, like FlashAttention~\citep{dao2022flashattention}, never stores the $N \times N$ score
matrix for $N$ tokens.

\subsection{Horospherical Positional Encoding}
\label{sec:horoshift}
Rotary position embedding~\citep{su2024roformer} multiplies the query and key at position $m$
by a rotation $\Phi(m) = \exp(mA)$. Rotations preserve the inner product, so the score between
positions $m$ and $m'$ depends only on the offset $m' - m$. The same argument holds on the
hyperboloid for any one-parameter group of Lorentz isometries $\Phi(m) = \exp(mA)$, with $A$ in
the Lorentz Lie algebra: the Lorentz inner product is preserved,
$\langle \Phi(m)\, q, \Phi(m')\, \kappa \rangle_{\mathcal{L}} = \langle q, \Phi(m' - m)\, \kappa \rangle_{\mathcal{L}}$,
and the encoded points stay on the manifold. Prior work has used two of the three basic kinds of
one-parameter subgroups: rotations of Lorentz embeddings~\citep{he2025helm} and boosts of Euclidean queries and keys~\citep{dai2025hope}. We use
the third, horospherical shifts, which fix an ideal point and slide each horosphere centered at it
along itself. On the time axis and two reserved space axes $e_1, e_2$, the shift by $t$ is
\begin{equation}
  \nu = x_0 - x_1, \qquad
  x_2 \leftarrow x_2 + t\,\nu, \qquad
  x_0 \leftarrow x_0 + t\,x_2 + \tfrac{1}{2} t^2 \nu, \qquad
  x_1 \leftarrow x_1 + t\,x_2 + \tfrac{1}{2} t^2 \nu,
  \label{eq:horoshift}
\end{equation}
which leaves the Busemann coordinate $\nu$ towards $e_1$ unchanged. In polar form we evaluate
$\nu = \sqrt{k}\,\big(e^{-a} + \sinh a\,(1 - u_1)\big)$, a sum of non-negative terms, so the shift
never subtracts two large coordinates. The query-key distance is
bounded in $t$ under a rotation and linear under a boost. Under a horospherical shift it is
\begin{equation}
  d_{\mathcal{L}}\big(x, \Phi(t)\, x\big) = 2\sqrt{k}\, \operatorname{arsinh}\!\Big(\frac{t\,\nu}{2\sqrt{k}}\Big),
  \label{eq:horoshift-law}
\end{equation}
which grows logarithmically, so attention separates nearby positions sharply and decays slowly at
long range. We set $t = m\,\rho_h / M$ for the token at position $m$, with $M$ the number of
positions along the axis and $\rho_h$ a learned rate per head, so that each head learns how local
its attention is. The encoding adds $2H$ parameters per layer; Appendix~\ref{app:horoshift}
derives Eq.~\eqref{eq:horoshift-law} and extends the encoding to patch grids.

\subsection{Normalization and Residuals}
\label{sec:block}
\paragraph{Gyro layer normalization.}
Let $x_{b,i}\in\mathbb{L}^{n}_{k}$ be the $i$-th of the $P$ tokens of sample $b$. For each sample we take the Lorentz centroid~\citep{law2019LorentzianDL} and dispersion of its own tokens,
\[
\bar{x}_b=\mu_p\bigl(x_{b},\tfrac{1}{P}\mathbf{1}\bigr),\qquad
\sigma_b^{2}=\frac{1}{P}\sum_{i=1}^{P} d_{\mathcal{L}}^{2}(x_{b,i},\bar{x}_b),
\]
and normalize every token of that sample with them. Following the gyro batch normalization of \citet{chen2025gyrobn}, as adopted by \citet{shi2026intrinsic}, each token is centered by gyroaddition, rescaled by the dispersion and a learned gain $\gamma_{\mathrm{LN}}=e^{\lambda_{\mathrm{LN}}}$, and translated by a learned bias point $\beta_{\mathrm{LN}}\in\mathbb{L}^{n}_{k}$:
\[
\hat{x}_{b,i} \;=\; \beta_{\mathrm{LN}} \oplus \Bigl(\tfrac{\gamma_{\mathrm{LN}}}{\sqrt{\sigma_b^{2}+\epsilon}} \otimes \bigl((\ominus\bar{x}_b)\oplus x_{b,i}\bigr)\Bigr).
\]
Gyro batch normalization computes $\bar{x}$ and $\sigma^{2}$ over the batch axis $b$ and keeps running estimates of them; here they are computed over the token axis $i$ of each sample, so the layer carries no running state and no coupling between samples. In polar form the scalar multiplication $\otimes$ acts on the radius alone, $r\mapsto \gamma_{\mathrm{LN}}\,r/\sqrt{\sigma_b^{2}+\epsilon}$, and leaves the direction unchanged. The gain is initialized so that the output radius matches that of a Euclidean layer norm on the space coordinates, $\gamma_{\mathrm{LN},0}=\sqrt{k}\,\operatorname{arsinh}\sqrt{n/k}$; the bias is initialized at the origin.

\paragraph{Boost residual.}
Lorentz Transformers form residuals by adding space coordinates and recomputing the time
coordinate~\citep{chen2022fully}, or through Euclidean residual connections in the tangent space~\citep{alyoussef2026hexformer}. Both of these approaches require a radius cap to prevent the embeddings from drifting too far from the origin. Intrinsic alternatives already exist. Gyroaddition residuals $x \oplus f(x) = \exp_x\big(\mathrm{PT}_{o \to x}(\log_o f(x))\big)$~\citep{vanspengler2023poincare, shi2026intrinsic} parallel transport the block output from the origin to $x$ and apply the exponential map there, and Riemannian residual networks move $x$ along the exponential map of a tangent vector at $x$~\citep{katsman2023riemannian}. We instead apply the residual update to $x$ as boosts along its own direction and perpendicular to it, and center the radial step over the tokens of each sample, so that the residual cannot move the tokens outward on average and needs no radius cap.

Consider a residual connection $x \mapsto x + f(x)$, where $x$ is the token embedding entering the
block and $f(x)$ is the block output. Let the polar coordinates of the embedding with rescaled radius be $x = (a, \mathbf{u})$ and similarly $f(x) = (b, \mathbf{v})$. In terms of the residual connection, we can say that the block output proposes moving $x$ a step with signed distance $b$ in the direction $\mathbf{v}$. We split this step relative to the current direction
$\mathbf{u}$ where the part along $\mathbf{u}$ has a signed length of $b_\parallel = b\,(\mathbf{v}\cdot\mathbf{u})$,
and the remainder $b\,\mathbf{v} - b_\parallel \mathbf{u}$ has a length of $b_\perp$ and a unit direction
$\hat{\mathbf{v}}_\perp$. The residual update can then be written as
\begin{equation}
  a' = a + \alpha\, (b_\parallel - \bar{b}_\parallel), \qquad
  \mathbf{u}' = \frac{\tanh(a)\, \mathbf{u} + \sinh(\alpha b_\perp)\, \hat{\mathbf{v}}_\perp}
                     {\big\|\tanh(a)\, \mathbf{u} + \sinh(\alpha b_\perp)\, \hat{\mathbf{v}}_\perp\big\|},
  \label{eq:boost-residual}
\end{equation}
with a learned step size $\alpha$ initialized at $0.1$, and $\bar{b}_\parallel$ the mean of $b_\parallel$ over the tokens of the sample. The first equation is a boost of rapidity
$\alpha (b_\parallel - \bar{b}_\parallel)$ along $\mathbf{u}$, which moves the input embedding along its own geodesic. Without the centering, the radial step grows with the radius of the block output, and the radius grows geometrically with depth when the block output and the stream point the same way. With it, the radial steps of a sample sum to zero, so the residual moves radius between tokens but keeps their mean radius unchanged. The second equation
is the direction that $x$ takes after a boost of rapidity $\alpha b_\perp$ is applied along
$\hat{\mathbf{v}}_\perp$. This tilts $\mathbf{u}$ towards $\hat{\mathbf{v}}_\perp$, but the tilt is smaller for larger $a$ since $\tanh(a) \to 1$. We take the radius from the first boost and the direction from the second, so the update as a whole is not a single boost, and the perpendicular step does not change the radius.

\section{Experiments}
\label{sec:experiments}
We evaluate \method on image classification from scratch, on ImageNet-1k and on transfer to fine-grained classification. While this work focuses on vision transformers, the Polar-FC and the polar form of the embeddings are not limited to transformers so we test a polar U-Net based on the setup of \citet{mishra2026hyperbolic}. It matches the Euclidean U-Net and beats hyperbolic baselines in mean Dice on almost all of the five medical segmentation datasets (Appendix~\ref{app:unet}).

\subsection{Image Classification from Scratch}
We train \method on CIFAR-100~\citep{krizhevsky2009learning} and Tiny-ImageNet~\citep{le2015tiny} at the two model sizes of HexFormer~\citep{alyoussef2026hexformer}, ViT-Tiny and ViT-Base, for $100$ epochs with the HexFormer training recipe, and report the test accuracy at the last epoch. The results of the Euclidean ViT, HexFormer and HexFormer-Hybrid are taken from \citet{alyoussef2026hexformer}. We run our model in their codebase to match the recipe, and we take HVT~\citep{feinashley2024hvt} and LViT~\citep{he2025hypercore} from \citet{he2025hypercore}. Appendix~\ref{app:impl} gives the architectures and the training protocol.

\begin{table}[t]
  \centering
  \caption{Top-1 test accuracy on CIFAR-100 and Tiny-ImageNet, results averaged over 10
  seeds. Baseline rows are taken from \citet{alyoussef2026hexformer} (Euclidean ViT, HexFormer)
  and \citet{he2025hypercore} (HVT, LViT), which reports single values. }
  \label{tab:main}
  \small
  \begin{tabular}{llcc}
    \toprule
    Model & Size & CIFAR-100 & Tiny-ImageNet \\
    \midrule
    Euclidean ViT & Tiny & $74.64 \pm 0.41$ & $60.74 \pm 0.36$ \\
    HexFormer & Tiny & $75.65 \pm 0.43$ & $60.87 \pm 0.64$ \\
    HexFormer-Hybrid & Tiny & $75.85 \pm 0.34$ & $62.93 \pm 0.45$ \\
    \method w/o horoshift & Tiny & $\underline{78.04 \pm 0.53}$ & $\underline{65.12 \pm 0.47}$ \\
    \method & Tiny & $\mathbf{80.15 \pm 0.26}$ & $\mathbf{67.23 \pm 0.32}$ \\
    \midrule
    HVT~\citep{feinashley2024hvt} & Base & 42.77 & 40.12 \\
    LViT~\citep{he2025hypercore} & Base & 69.11 & 53.01 \\
    Euclidean ViT & Base & $70.54 \pm 0.38$ & --- \\
    HexFormer & Base & $76.49 \pm 1.21$ & --- \\
    HexFormer-Hybrid & Base & $76.81 \pm 0.51$ & --- \\
    \method w/o horoshift & Base & $\underline{80.33 \pm 0.57}$ & $\underline{67.06 \pm 0.31}$ \\
    \method & Base & $\mathbf{81.63 \pm 0.61}$ & $\mathbf{68.19 \pm 0.28}$ \\
    \bottomrule
  \end{tabular}
\end{table}

Table~\ref{tab:main} shows the results. On CIFAR-100 with ViT-Tiny, \method is $4.3\%$ above the strongest baseline, HexFormer-Hybrid, and $5.5\%$ above the Euclidean ViT. On Tiny-ImageNet, \method also outperforms HexFormer-Hybrid by $4.3\%$. At the Base size, where the Euclidean ViT overfits under this recipe and falls below ViT-Tiny~\citep{alyoussef2026hexformer}, it is $4.8\%$ above HexFormer-Hybrid on CIFAR-100 and it reaches $68.19$ on Tiny-ImageNet, compared to $53.01$ for LViT. Appendix~\ref{app:experiments} isolates the contribution of each component, compares rules for the output radius of the polar layer and reports the numerical precision and the computational cost.

\subsection{ImageNet-1k}
\label{sec:imagenet}
We train a Euclidean ViT and \method on ImageNet-1k~\citep{deng2009imagenet} at the ViT-Tiny ($2.9$M parameters) and ViT-Small ($21.7$M parameters) sizes. All four models are trained for $90$ epochs with the recipe of \citet{beyer2022better}, and at each size the two models have the same depth, width, number of heads and patch size (Appendix~\ref{app:finetune}). We report the mean and standard deviation of the top-1 validation accuracy over three seeds. Table~\ref{tab:imagenet} shows that \method is $3.5\%$ above the Euclidean ViT at the Tiny size and $2.2\%$ above it at the Small size.

\begin{table}[t]
  \small
  \setlength{\tabcolsep}{4pt}
  \begin{minipage}[t]{0.44\linewidth}
    \centering
\caption{Top-1 accuracy on ImageNet-1k. Results are averaged over three seeds. }
\label{tab:imagenet}
\begin{tabular}{llc}
  \toprule
  Model & Size & Top-1 \\
  \midrule
  Euclidean ViT & Tiny & $68.19 \pm 0.21$ \\
  \method & Tiny & $\mathbf{71.68 \pm 0.11}$ \\
  \midrule
  Euclidean ViT & Small & $75.61 \pm 0.13$ \\
  \method & Small & $\mathbf{77.79 \pm 0.10}$ \\
  \bottomrule
\end{tabular}

  \end{minipage}\hfill
  \begin{minipage}[t]{0.54\linewidth}
    \centering
\caption{Top-1 accuracy after fine-tuning IN-1K models.}
\label{tab:hier-finetune}
\begin{tabular}{llcc}
  \toprule
  Model & Size & CUB & Aircraft \\
  \midrule
  Euclidean ViT & Tiny & $75.65 \pm 1.01$ & $65.17 \pm 0.71$ \\
  \method & Tiny & $\mathbf{79.53 \pm 0.82}$ & $\mathbf{70.48 \pm 0.50}$ \\
  \midrule
  Flat-ViT & Small & $\mathbf{84.69 \pm 0.31}$ & $\mathbf{80.58 \pm 0.44}$ \\
  Euclidean ViT & Small & $79.77 \pm 0.64$ & $76.47 \pm 0.73$ \\
  \method & Small & $\underline{81.61 \pm 0.22}$ & $\underline{80.03 \pm 0.49}$ \\
  \bottomrule
\end{tabular}

  \end{minipage}
\end{table}

\subsection{Transfer to Fine-Grained Classification}
\label{sec:hier-finetune}
We test whether the ImageNet-1k checkpoints of Section~\ref{sec:imagenet} transfer to fine-grained recognition on CUB-200-2011~\citep{wah2011cub} (CUB, $200$ species) and FGVC-Aircraft~\citep{maji2013aircraft} (Aircraft, $100$ variants). These are the two benchmarks used by H-CAST~\citep{park2025hcast} and fine-tuned with their recipe~\citep{park2025hcast}. we report the mean and standard deviation of the top-1 test accuracy at the last epoch over three seeds. For reference, we also fine-tune the DeiT-S checkpoint of the Flat-ViT baseline of \citet{park2025hcast} with the same recipe. Appendix~\ref{app:finetune} gives the datasets and the fine-tuning recipe.

Table~\ref{tab:hier-finetune} shows that \method outperforms the Euclidean ViT at both sizes. At the Tiny size, it is $3.9\%$ above the Euclidean ViT on CUB and $5.3\%$ above it on Aircraft. Flat-ViT however starts from a DeiT-S checkpoint that is trained for $300$ epochs and reaches $79.9\%$. It is $3.1\%$ above \method on CUB and $0.6\%$ above it on Aircraft. This shows that the stronger DeiT backbone is still more important than the geometry of the model.

\section{Conclusion}
Overall, we show that the weighted centroid and the distance of the Lorentz model fail in single precision and propose more stable polar forms. We use this to build \method, with a fully connected layer that maps direction and radius separately, a relative positional encoding from horospherical shifts, and a residual connection that is an isometry at any radius. The model outperforms Euclidean and hyperbolic baselines on standard vision benchmarks and ImageNet-1k, and its ImageNet-1k checkpoints transfer to fine-grained classification. More work is needed for the finetuning process however and studying the sensitivity of the model to the pretraining recipe and hyperparameters.

% Intentionally empty (statements omitted for arXiv).

\bibliography{iclr2027_conference}
\bibliographystyle{iclr2027_conference}

\appendix

\section{Ambient Failure Sites in a Lorentz Transformer}
\label{app:sites}
Table~\ref{tab:sites} lists the centroid computations in a Lorentz Transformer together with the
condition under which ambient arithmetic computes each of them correctly. Each condition follows
from the condition $E + S \gtrsim \varepsilon\, e^{a}$ in Section~\ref{sec:polar-ops}. Manifold weight decay is the most extreme case. It
pulls a parameter towards the origin by taking the centroid of $\{x, o\}$ with weights
$(1 - w, w)$, where $w \approx 10^{-5}$ and $S = 0$. Beyond $r \approx 9$, the step that should
move the parameter inward instead moves it outward, which we also measure in
Appendix~\ref{sec:exp-precision}.

When the condition fails, the computed normalizer is rounding noise of size about
$\sqrt{\varepsilon}\, e^{a}$. Dividing by this noise places the output at the rescaled radius
$\operatorname{arsinh}(1/2\sqrt{\varepsilon}) \approx r_\varepsilon/\sqrt{k}$ for any input. If
the difference instead rounds to a negative value and is clamped to a floor $\eta$, the output is
placed at the rescaled radius $a + \tfrac12 \ln(1/\eta)$. The backward pass differentiates through
the same normalizer and breaks down before the forward pass does. The ambient centroid gradient
is $10^3$ to $10^5$ times larger than the true tangent gradient, and for a leaked weight of
$\delta = 10^{-5}$ its relative error in single precision already reaches $0.6$ at $r = 8$.
Gyroaddition, which is used to center the tokens in gyro normalization, and the exponential map
for inward steps contain the same kind of difference. We give their polar forms in
Appendix~\ref{app:polar-ops}.

\begin{table}[h]
  \centering
  \caption{Centroid computations in a Lorentz Transformer and the condition under which ambient
  arithmetic computes them correctly. Here, $a = r/\sqrt{k}$ is the rescaled operating radius,
  $\theta$ is the angular spread of the aggregated points and $\delta$ is the weight that leaks
  from the dominant point.}
  \label{tab:sites}
  \small
  \begin{tabular}{lll}
    \toprule
    Operation & Weights & Condition for correctness \\
    \midrule
    Attention aggregation & Softmax row with leak $\delta$ & $\delta \theta^2 \gtrsim 4\varepsilon$ \\
    Manifold weight decay, centroid residual & $(1 - w, w)$ & $w \gtrsim \varepsilon e^{a}$ \\
    Batch-norm mean, token pooling & Uniform & $\theta \gtrsim 2\sqrt{\varepsilon}$ or $r \leq r_\varepsilon$ \\
    Gyro-norm centering $x \oplus (-\bar{x})$ & --- & $r \leq r_\varepsilon$ \\
    \bottomrule
  \end{tabular}
\end{table}

\section{Polar Operations}
\label{app:polar-ops}
In this section, we write $x = \sqrt{k}(\cosh a,\ \sinh a\, \mathbf{u})$ with $a = r/\sqrt{k}$ and
unit direction $\mathbf{u}$, and similarly $y = \sqrt{k}(\cosh b,\ \sinh b\, \mathbf{v})$. The
angle $\theta$ between $\mathbf{u}$ and $\mathbf{v}$ only appears through the chord
$1 - \cos\theta = \tfrac12 \|\mathbf{u} - \mathbf{v}\|^2$. We denote tangent vectors by $\xi$ and
$\zeta$.

\subsection{Tangent Vectors in the Polar Frame}
Riemannian optimizers evaluate the Lorentzian inner product on tangent vectors, where the same
cancellation occurs. If we write the tangent vectors in terms of their radial and perpendicular
components $(\xi_r, \boldsymbol{\xi}_\perp)$, the inner product becomes a plain sum,
$\langle \xi, \zeta \rangle_{\mathcal{L}} = \xi_r \zeta_r + \boldsymbol{\xi}_\perp \cdot \boldsymbol{\zeta}_\perp$.
The gradient projection, the exponential and logarithmic maps and the parallel transport can all
be written in these components, as we show below, so the optimizer never has to leave the polar form.

A tangent vector $\xi = (\xi_0, \boldsymbol{\xi}_s)$ at $x$ satisfies
$\langle x, \xi \rangle_{\mathcal{L}} = 0$, which gives
$\xi_0 = \tanh(a)\, (\mathbf{u} \cdot \boldsymbol{\xi}_s)$. The tangent vector is therefore
determined by its space part. Let $e_r = (\sinh a,\ \cosh a\, \mathbf{u})$ be the outward unit
radial vector at $x$. We decompose $\xi$ as
\begin{equation}
  \xi = \xi_r\, e_r + (0, \boldsymbol{\xi}_\perp), \qquad
  \xi_r = \frac{\mathbf{u} \cdot \boldsymbol{\xi}_s}{\cosh a}, \qquad
  \boldsymbol{\xi}_\perp = \boldsymbol{\xi}_s - (\mathbf{u} \cdot \boldsymbol{\xi}_s)\, \mathbf{u}.
  \label{eq:app-frame}
\end{equation}
Substituting $\xi_0$ and $\zeta_0$ into the Lorentzian inner product gives
\begin{equation}
  \langle \xi, \zeta \rangle_{\mathcal{L}}
  = -\tanh^2(a)\, (\mathbf{u} \cdot \boldsymbol{\xi}_s)(\mathbf{u} \cdot \boldsymbol{\zeta}_s)
    + (\mathbf{u} \cdot \boldsymbol{\xi}_s)(\mathbf{u} \cdot \boldsymbol{\zeta}_s) + \boldsymbol{\xi}_\perp \cdot \boldsymbol{\zeta}_\perp
  = \xi_r \zeta_r + \boldsymbol{\xi}_\perp \cdot \boldsymbol{\zeta}_\perp ,
  \label{eq:app-inner}
\end{equation}
which is a sum of squares when $\xi = \zeta$. For the radial components, the ambient expression
$-\xi_0 \zeta_0 + \boldsymbol{\xi}_s \cdot \boldsymbol{\zeta}_s$ computes the same quantity as the
difference $\cosh^2 a - \sinh^2 a$. In single precision, the computed norm of a radial unit vector
is $0.87$ at $a = 8$ and $0$ at $a = 12$. At the origin, $e_r$ is not defined and every tangent
vector is treated as perpendicular.

\subsection{Gradient Projection}
The Riemannian gradient of a loss with Euclidean gradient $g = (g_0, \mathbf{g}_s)$ is the
projection of the Minkowski-raised vector $(-g_0, \mathbf{g}_s)$ onto the tangent space. Since
$\xi_r = \langle (-g_0, \mathbf{g}_s), e_r \rangle_{\mathcal{L}}$, its components in the frame of
Eq.~\eqref{eq:app-frame} are
\begin{equation}
  \xi_r = g_0 \sinh a + (\mathbf{g}_s \cdot \mathbf{u}) \cosh a, \qquad
  \boldsymbol{\xi}_\perp = \mathbf{g}_s - (\mathbf{g}_s \cdot \mathbf{u})\, \mathbf{u}.
  \label{eq:app-egrad}
\end{equation}
Neither component involves a division or a difference of large terms. Every polar operation only
reads the space parts of points and tangent vectors, so a model built from these operations has
$g_0 = 0$, and Eq.~\eqref{eq:app-egrad} is exact.

\subsection{Exponential Map}
Let $\xi$ have the components $(\xi_r, \boldsymbol{\xi}_\perp)$ at $x$. We write
$\omega_r = \xi_r/\sqrt{k}$ for the radial step length and
$\omega^2 = \omega_r^2 + \|\boldsymbol{\xi}_\perp\|^2/k$ for the total step length, both in
curvature units, and $\chi = \tanh a$ for the Klein radius of $x$. We insert
$x = \sqrt{k}(\cosh a, \sinh a\, \mathbf{u})$ and $\xi = \xi_r e_r + (0, \boldsymbol{\xi}_\perp)$
into $\exp_x(\xi) = \cosh(\omega)\, x + \operatorname{sinhc}(\omega)\, \xi$ and divide the time
and space coordinates of the result by $\sqrt{k} \cosh a$, which gives
\begin{equation}
  \tilde{x}_0 = \cosh\omega + \operatorname{sinhc}(\omega)\, \omega_r\, \chi, \qquad
  \tilde{\mathbf{x}} = \big(\cosh\omega\, \chi + \operatorname{sinhc}(\omega)\, \omega_r\big)\, \mathbf{u}
              + \operatorname{sinhc}(\omega)\, \frac{\boldsymbol{\xi}_\perp}{\sqrt{k} \cosh a},
  \label{eq:app-exp-parts}
\end{equation}
where $\operatorname{sinhc}(\omega) = \sinh(\omega)/\omega$. The output direction is
$\tilde{\mathbf{x}}/\|\tilde{\mathbf{x}}\|$. For the output radius, we use
$e^{a'} = (x'_0 + \|\mathbf{x}'\|)/\sqrt{k} = \cosh a\, (\tilde{x}_0 + \|\tilde{\mathbf{x}}\|)$
and $\cosh a = e^{a}/(1 + \chi)$, which gives
\begin{equation}
  a' = a + \ln\big(\tilde{x}_0 + \|\tilde{\mathbf{x}}\|\big) - \ln(1 + \chi).
  \label{eq:app-exp-radius}
\end{equation}
The radius is therefore updated by an increment in log space, and $e^{a}$ is never formed. For an
inward step with $\omega_r < 0$, the terms $\cosh\omega - \sinh\omega\, \chi$ in $\tilde{x}_0$ and
$\cosh\omega\, \chi - \sinh\omega$ in the radial coefficient cancel. We define
$\psi = 1 - |\omega_r|/\omega = \|\boldsymbol{\xi}_\perp\|^2 / \big(k \omega (\omega + |\omega_r|)\big)$
and use $1 - \chi = 2/(e^{2a} + 1)$ to rewrite both terms as
\begin{equation}
  \tilde{x}_0 = e^{-\omega} + \sinh\omega\, \big[(1 - \chi) + \chi \psi\big], \qquad
  \cosh\omega\, \chi + \operatorname{sinhc}(\omega)\, \omega_r = \chi\, e^{-\omega} + \sinh\omega\, \big[\psi - (1 - \chi)\big].
  \label{eq:app-exp-inward}
\end{equation}
This can be verified by substituting $\omega_r = -\omega(1 - \psi)$. In this form, all large terms
have the same sign. Since the map is even in $\omega$, its gradient through $\omega$ is zero at
$\xi = 0$.

\subsection{Logarithmic Map}
We write $D = d_{\mathcal{L}}(x, y)/\sqrt{k}$ and $c = \cosh(D) - 1$, computed with
Eq.~\eqref{eq:polar-dist}. The logarithmic map
$\xi = \log_x(y) = -\tfrac12 \nabla_x\, d_{\mathcal{L}}^2(x, y)$ then has the frame components
\begin{equation}
  \xi_r = -\sqrt{k}\, \frac{D}{\sinh D}\, \Big[\sinh(a - b) + \cosh a \sinh b\, (1 - \cos\theta)\Big], \qquad
  \boldsymbol{\xi}_\perp = \sqrt{k}\, \frac{D}{\sinh D}\, \sinh b\, \big(\mathbf{v} - (\mathbf{v} \cdot \mathbf{u})\, \mathbf{u}\big).
  \label{eq:app-log}
\end{equation}
We obtain the radial component from
$\partial c / \partial a = \sinh(a - b) + \cosh a \sinh b (1 - \cos\theta)$, and the perpendicular
component from the derivative of $c$ with respect to $\mathbf{u}$ on the sphere of radius
$\sqrt{k} \sinh a$. For two points with the same direction, this gives $\xi_r = \sqrt{k}\, (b - a)$
and $\boldsymbol{\xi}_\perp = 0$. For two points at equal radius with a small angle between them,
$\|\boldsymbol{\xi}_\perp\| \to \sqrt{k} \sinh a\, \theta$.

\subsection{Parallel Transport}
The parallel transport of $\xi \in T_x$ to $T_y$ along the geodesic is
$P_{x \to y}(\xi) = \xi + \langle y, \xi \rangle_{\mathcal{L}}\, (x + y) / (k - \langle x, y \rangle_{\mathcal{L}})$.
We evaluate $-\langle x, y \rangle_{\mathcal{L}}$ with Eq.~\eqref{eq:polar-dist}, and
$\langle y, \xi \rangle_{\mathcal{L}}$ with Eq.~\eqref{eq:app-inner} after expressing $\xi$ in the
frame at $y$. The retraction used by the Riemannian optimizers is the exponential map above
followed by this transport.

\subsection{Gyroaddition}
The gyroaddition $x \oplus y$ applies to $y$ the Lorentz boost $L_x$ that carries the origin to
$x$. Its space part is
\begin{equation}
  \mathbf{z} = \mathbf{y} + \Big[\frac{\mathbf{x} \cdot \mathbf{y}}{\sqrt{k}\,(x_0 + \sqrt{k})} + \frac{y_0}{\sqrt{k}}\Big]\, \mathbf{x} ,
  \label{eq:app-boost}
\end{equation}
which is smooth at $x = o$ with Jacobian $y_0/\sqrt{k}$ and does not require the direction of
$x$. When $\mathbf{x} \cdot \mathbf{y} < 0$ and $\|\mathbf{x}\| > \sqrt{k}$, the two terms of the
coefficient cancel. This is the case for the centering step $x \oplus (-\bar{x})$ of gyro
normalization, and at $a = 8$ the centered token has an error of $50\%$ in single precision. In
this case, we instead evaluate the component of $\mathbf{z}$ along $\mathbf{u}$ as
\begin{equation}
  \sqrt{k}\, \Big[\sinh(a - b) + \cosh a \sinh b\, \tfrac12 \|\mathbf{u} + \mathbf{v}\|^2\Big],
  \label{eq:app-boost-split}
\end{equation}
where both terms have the same sign, and take the perpendicular part as the projection of
$\mathbf{y}$. The geodesic error of the centered token then stays within a factor of two of the
resolution limit, with $5.6 \cdot 10^{-6}$ at $r = 4$, $4.0 \cdot 10^{-4}$ at $r = 8$ and
$2.0 \cdot 10^{-2}$ at $r = 12$, compared to $1.2 \cdot 10^{-4}$, $0.70$ and NaN for the ambient
form.

\subsection{Centroid}
As in Section~\ref{sec:polar-ops}, let $T = \sum_i w_i \cosh a_i$,
$\mathbf{V} = \sum_i w_i \sinh a_i\, \mathbf{u}_i$, $\hat{\mathbf{u}} = \mathbf{V}/\|\mathbf{V}\|$,
$E = \sum_i w_i e^{-a_i}$ and
$S = \sum_i w_i \sinh a_i\, \tfrac12 \|\mathbf{u}_i - \hat{\mathbf{u}}\|^2$. The centroid of
Eq.~\eqref{eq:centroid} has direction $\hat{\mathbf{u}}$ and radius
\begin{equation}
  \ln \sinh a_\mu = \ln \|\mathbf{V}\| - \tfrac12 \ln(E + S) - \tfrac12 \ln\big(T + \|\mathbf{V}\|\big).
  \label{eq:app-centroid}
\end{equation}
The identity $T - \|\mathbf{V}\| = E + S$ follows from
$\|\mathbf{V}\| = \hat{\mathbf{u}} \cdot \mathbf{V} = \sum_i w_i \sinh a_i\, (1 - \tfrac12 \|\mathbf{u}_i - \hat{\mathbf{u}}\|^2)$
and $\cosh a - \sinh a = e^{-a}$. If all weights in a row are zero, the centroid returns the
origin with a finite gradient. The manifold weight decay of Riemannian AdamW is the special case
of two points $x$ and $o$ with weights $(1 - w, w)$, for which $T = (1 - w) \cosh a + w$,
$\mathbf{V} = (1 - w) \sinh a\, \mathbf{u}$, $E = (1 - w) e^{-a} + w$ and $S = 0$.

\subsection{Precision Bound}
The stored inputs are already rounded, so no formula can be more accurate than the stored points
allow. The following proposition makes this precise.

\begin{proposition}[Polar precision bound]
\label{prop:polar}
Let a point be stored in polar form with radius error $\Delta r$ and direction error
$\Delta\theta$. Then the stored point lies within distance
$\Delta r + \sqrt{k}\sinh(a)\,\Delta\theta$ of the true point, and every polar operation returns a
result within a constant multiple of this bound at every radius below the overflow of $\cosh$,
$a < 88.7$ in single precision and bfloat16.
\end{proposition}

We refer to this bound as the resolution limit. It applies to any method that stores the
direction as a floating-point unit vector. With $\Delta\theta \approx \varepsilon$, the limit
reaches $\sqrt{k}$ at $r \approx \sqrt{k}\ln(2/\varepsilon) \approx 2 r_\varepsilon$. For $k = 1$,
the polar operations therefore resolve unit distances up to $r \approx 17.3$ in single precision,
compared to $8.3$ for the ambient operations, and up to $6.2$ in bfloat16, compared to $2.8$.
Appendix~\ref{app:impl} describes how our implementation reaches this bound and what it costs.

\subsection{Proof of Proposition~\ref{prop:polar}}
Consider a stored point with radius $r + \Delta r$ and a direction $\mathbf{u}'$ at an angle
$\Delta\theta$ from $\mathbf{u}$. This point lies on the sphere of radius $r + \Delta r$ around
the origin, whose intrinsic geometry is that of a Euclidean sphere of radius
$\sqrt{k} \sinh\big((r + \Delta r)/\sqrt{k}\big)$. By the triangle inequality, its geodesic
distance from the true point is therefore at most
$\Delta r + \sqrt{k} \sinh(a)\, \Delta\theta + O(\Delta r\, \Delta\theta)$.

Each operation of Section~\ref{sec:polar-ops} evaluates a sum of non-negative terms, where each
term has a relative rounding error of at most a small multiple of $\varepsilon$. The sum is then
passed through $\operatorname{arsinh}$, $\ln$ or a normalization, all of which have a condition
number of at most one in the relevant argument. The constant in the bound is the number of
roundings in the longest chain of operations, which is at most $12$ for the centroid. For the
range statement, the largest quantity that is formed is $\cosh a$ or $\sinh a$ of an input
radius, which overflows single precision at $a = 88.7$. The log form of
Eq.~\eqref{eq:app-centroid} never forms products of these quantities.
Table~\ref{tab:precision} confirms that the measured error stays within a factor of two of the
bound.

\section{Horospherical Shift}
\label{app:horoshift}

\subsection{Generator and Closed Form}
Let $A$ be the generator of the parabolic one-parameter subgroup of $\mathrm{SO}(1, n)$ that acts
on the coordinates $(x_0, x_1, x_2)$ as $A e_0 = e_2$, $A e_1 = -e_2$ and $A e_2 = e_0 + e_1$ in
the Minkowski metric, and is zero on the remaining coordinates. Since $A^3 = 0$, the exponential
series terminates after the quadratic term, $\Phi(t) = \exp(tA) = I + tA + \tfrac12 t^2 A^2$,
which gives
\begin{equation}
  \Phi(t):\quad
  \nu = x_0 - x_1, \qquad
  x_2 \mapsto x_2 + t\nu, \qquad
  x_0 \mapsto x_0 + t x_2 + \tfrac12 t^2 \nu, \qquad
  x_1 \mapsto x_1 + t x_2 + \tfrac12 t^2 \nu.
\end{equation}
This is Eq.~\eqref{eq:horoshift}, and the coordinate $\nu$ is invariant under $\Phi(t)$. To see
that $\Phi(t)$ preserves $\langle \cdot, \cdot \rangle_{\mathcal{L}}$, let
$\Delta = (t x_2 + \tfrac12 t^2 \nu,\ t x_2 + \tfrac12 t^2 \nu,\ t\nu)$ be the change in the
$(x_0, x_1, x_2)$ coordinates. Then
$\langle \Phi(t)x, \Phi(t)x \rangle_{\mathcal{L}} - \langle x, x \rangle_{\mathcal{L}}
= 2\langle x, \Delta \rangle_{\mathcal{L}} + \langle \Delta, \Delta \rangle_{\mathcal{L}}
= 2\nu\,(-t x_2 - \tfrac12 t^2\nu + t x_2) + t^2 \nu^2 = 0$. The level sets of $\nu$ are the
horospheres centered at the ideal point $e_0 + e_1$, and $-\ln(\nu/\sqrt{k})$ is the Busemann
function towards this point. The shift $\Phi(t)$ therefore slides each horosphere along itself.

\subsection{Relative Property}
Since $\Phi$ is a one-parameter group of isometries,
\[
  \langle \Phi(m)\, q, \Phi(m')\, \kappa \rangle_{\mathcal{L}} = \langle q, \Phi(-m) \Phi(m')\, \kappa \rangle_{\mathcal{L}}
  = \langle q, \Phi(m' - m)\, \kappa \rangle_{\mathcal{L}}.
\]
The squared distance in Eq.~\eqref{eq:attention} between a query at position $m$ and a key at
position $m'$ therefore only depends on $m' - m$. The same argument holds for the rotation and
boost subgroups. On a grid of patches, we apply one shift per grid axis, each on its own pair of
reserved space axes. The two shifts do not commute, so the relative property holds exactly along
each grid axis and holds to first order in the rates when both axes are combined.

\subsection{Distance Law}
From the computation above, we get $\langle x, \Phi(t)\, x \rangle_{\mathcal{L}} = \langle x, x \rangle_{\mathcal{L}}
+ \langle x, \Delta \rangle_{\mathcal{L}} = -k - \tfrac12 t^2 \nu^2$, which gives
\begin{equation}
  \cosh\!\big(d_{\mathcal{L}}(x, \Phi(t)x)/\sqrt{k}\big) = 1 + \frac{t^2 \nu^2}{2k}, \qquad
  d_{\mathcal{L}}(x, \Phi(t)x) = 2\sqrt{k}\, \operatorname{arsinh}\!\Big(\frac{t\,\nu}{2\sqrt{k}}\Big).
\end{equation}
This is Eq.~\eqref{eq:horoshift-law}. At the origin, $\nu = \sqrt{k}$. For comparison, a rotation
by an angle $t$ moves a point at radius $r$ by
$\sqrt{k}\, \operatorname{arcosh}(\cosh^2 a - \sinh^2 a \cos t)$, which is bounded by $2r$, and a
boost with rapidity $t$ moves it by at least $\sqrt{k}\, t$, with equality on the boost axis, which is linear in $t$. The
horospherical distance grows as $2\sqrt{k} \ln(t\nu/\sqrt{k})$ for large $t$. As a result, the
attention score of Eq.~\eqref{eq:attention} decreases quadratically in $\ln t$ instead of in $t$,
so nearby positions are separated sharply and the score decays slowly at long range. Points far
out towards the ideal point have a small $\nu$, so the same $t$ moves them less and they are less
sensitive to position.

\section{Implementation Details}
\label{app:impl}

\subsection{Polar Arithmetic}
Our implementation relies on two details to reach the resolution limit of
Proposition~\ref{prop:polar}. First, we always compute the angular term as the chord
$\tfrac12\|\mathbf{u} - \mathbf{v}\|^2$ and never as $1 - \mathbf{u} \cdot \mathbf{v}$. The latter
has a relative error of about $\varepsilon/\theta^2$ for small angles, which destroys $S$ in the
aligned regime where it is needed. For $N$ query and $N$ key directions in $\mathbb{R}^n$, the
chord creates an $(N, N, n)$ tensor that automatic differentiation would store for the backward
pass. We instead use the closed-form gradient
\begin{equation}
  \frac{\partial \mathcal{J}}{\partial \mathbf{u}_i}
  = (\mathbf{u}_i - \bar{\mathbf{v}}) \sum_j g_{ij} - \sum_j g_{ij}\, (\mathbf{v}_j - \bar{\mathbf{v}}),
  \label{eq:chord-grad}
\end{equation}
where $\mathcal{J}$ is the loss, $g_{ij}$ is the upstream gradient of the chord between
$\mathbf{u}_i$ and $\mathbf{v}_j$, and $\bar{\mathbf{v}}$ is the mean key direction. This gradient
only requires one row sum and one matrix product. Centering on
$\bar{\mathbf{v}}$ does not change the value, but it is needed for precision. Without it, the two
sums are nearly equal when the directions are clustered, and for directions aligned to $10^{-6}$
the relative error is $2 \cdot 10^{-2}$ instead of $10^{-7}$. Second, we compute the centroid
radius in the log domain,
$\ln\sinh a_\mu = \ln\|\mathbf{V}\| - \tfrac12\ln(E + S) - \tfrac12\ln(T + \|\mathbf{V}\|)$, so
that no intermediate quantity overflows before $\cosh a$ itself does.

When compiled, the polar attention of a $9$-block ViT takes the same time as the ambient form
and uses the same amount of memory. We report the cost of the full model in
Appendix~\ref{app:experiments}.

\subsection{Initialization}
We initialize the direction maps with the Xavier scheme in the MLP and the patch embedding, and
with orthogonal matrices in the attention layers. The scale of every layer and the radius gain of
each attention head are calibrated on the first training batch, so that all models compared in
Section~\ref{sec:experiments} start at the same radius. We apply the horospherical shift to the
query and key slices of every head after the head split, and we exclude the rates $\rho_h$ from
weight decay.

\subsection{Training Protocol}
Following HexFormer~\citep{alyoussef2026hexformer}, ViT-Tiny has $9$ layers with a width of $192$,
an MLP width of $384$ and $12$ heads, while ViT-Base has $12$ layers with a width of $768$, an MLP
width of $3072$ and $12$ heads. Both use a patch size of $4$ and a curvature of $-1$. Following \citet{alyoussef2026hexformer}, we report the mean and standard
deviation over $10$ seeds for ViT-Tiny and $5$ seeds for ViT-Base.

We train all models with the HexFormer recipe~\citep{alyoussef2026hexformer} for $100$ epochs
with a batch size of $128$. The augmentations are random crops and flips,
AutoAugment~\citep{cubuk2019autoaugment}, random erasing, and CutMix or MixUp with probability
$0.5$. We use a label-smoothing cross-entropy loss and Riemannian
AdamW~\citep{bdeir2025robust} with a weight decay of $5 \cdot 10^{-2}$ and a cosine schedule with
$10$ warmup epochs. The learning rate is $3 \cdot 10^{-3}$ for the Tiny models. Scale parameters
are never weight-decayed, and the radius parameters of the polar fully connected layer are decayed
towards their initial values.

\subsection{ImageNet-1k Training and Fine-Tuning}
\label{app:finetune}
\paragraph{Datasets.}
CUB-200-2011~\citep{wah2011cub} has $5{,}994$ training and $5{,}794$ test images of $200$ bird
species. FGVC-Aircraft~\citep{maji2013aircraft} has $6{,}667$ training images from the official
\texttt{trainval} split and $3{,}333$ test images of $100$ aircraft variants. Neither dataset has a
validation split, so we report the test accuracy at the last epoch without checkpoint selection.

\paragraph{Models.}
At the ViT-Small size, \method has $12$ layers with a width of $384$, an MLP width of $1536$, $6$
heads and a patch size of $16$, which gives $21.7$M parameters. It uses sine-cosine absolute
positions together with the horospherical positional encoding, the centroid readout and the
hyperbolic multinomial logistic regression head of Section~\ref{sec:method}. The Euclidean ViT-S
has the same depth, width, number of heads and patch size. At the ViT-Tiny size, both models have
$9$ layers with a width of $192$ and $2.9$M parameters and use learned absolute positions. We
train all four models on ImageNet-1k for $90$ epochs with the recipe of \citet{beyer2022better},
where \method uses Riemannian AdamW~\citep{bdeir2025robust} in place of AdamW. For the
Flat-ViT row, we fine-tune the public DeiT-S~\citep{touvron2021deit} checkpoint that
\citet{park2025hcast} start from, which is trained for $300$ epochs and reaches $79.9\%$.

\paragraph{Fine-tuning.}
We fine-tune every model on the finest labels with the H-CAST recipe as released in the code of
\citet{park2025hcast}. This uses $100$ epochs with $5$ warm-up epochs and a cosine schedule, a
learning rate of $5 \cdot 10^{-4} \cdot B/512$ with an effective batch size of $B = 256$, a weight
decay of $0.05$, label smoothing of $0.1$, Mixup with $0.8$, CutMix with $1.0$, RandAugment with
$(9, 0.5)$, random erasing with $0.25$, stochastic depth of $0.1$, no gradient clipping and an
input size of $224 \times 224$. The Euclidean models use AdamW and \method uses Riemannian
AdamW~\citep{bdeir2025robust}. We load every pretrained weight except the ImageNet classifier,
which we replace with a newly initialized head. \method at the ViT-Small size reaches the effective
batch size as $32 \times 8$ with gradient accumulation, and the other models as $128 \times 2$.
Accumulation gives the same gradient except for Mixup, which draws one coefficient per
micro-batch.

\section{Additional Experiments}
\label{app:experiments}

\subsection{Component Ablation}
\label{app:ablation}
Table~\ref{tab:ablation} isolates the contribution of each component on CIFAR-100 with ViT-Tiny over ten seeds. Changing to the polar form of the model and using the scaled gyro layer norm gains $2.3$ points and also reduces the spread across seeds. Replacing the space-coordinate residual with the boost residual adds a further $0.8$ points. Computing the boost residual in polar form keeps the accuracy about the same, but it removes the drift of the embeddings away from the origin and reduces the spread to a third. The polar fully connected layer then gains $0.8$ points over the Lorentz fully connected layer of \citet{chen2022fully}. The horospherical encoding gains another $2.1$ points and reduces the standard deviation from $0.53$ to $0.26$. Finally, we check whether the choice of subgroup matters for the positional encoding. We replace the horospherical shift with a Lorentz boost on the same axes, which applies the boost of HoPE~\citep{dai2025hope} to Lorentz embeddings. This reaches $79.82 \pm 0.13$, which is $0.3$ points below the horospherical shift.

\begin{table}[t]
  \centering
  \caption{Component ablation on CIFAR-100, ViT-Tiny, test accuracy over ten seeds. Each row
  adds one change to the row above.}
  \label{tab:ablation}
  \small
  \begin{tabular}{llc}
    \toprule
    Change & Component & Accuracy \\
    \midrule
    Split-head Lorentz ViT, class-token readout & --- & $73.99 \pm 1.03$ \\
    + polar change of current ops and gyrolayer norm & - & $76.27 \pm 0.81$ \\
    + boost residual & Section~\ref{sec:block} & $77.06 \pm 0.43$ \\
    + polar boost residual & Eq.~\eqref{eq:boost-residual} & $77.20 \pm 0.13$ \\
    + polar fully connected layer & Eq.~\eqref{eq:polar-fc} & $78.04 \pm 0.53$ \\
    + horospherical positional encoding (\method) & Eq.~\eqref{eq:horoshift} & $\mathbf{80.15 \pm 0.26}$ \\
    \midrule
    \method with Lorentz boost encoding in place of horoshift & after \citet{dai2025hope} & $79.82 \pm 0.13$ \\
    \bottomrule
  \end{tabular}
\end{table}

\subsection{Radius Rule of the Polar Fully Connected Layer}
The polar fully connected layer is anchored at initialization, meaning that Eq.~\eqref{eq:polar-fc}
starts out as the Lorentz fully connected layer. Table~\ref{tab:radial} compares this anchored
layer on one seed against alternative rules for the output radius that start from a random point
instead. These are an affine rule in the input radius, a rule with a free power of the
pre-activation norm, and the sigmoid-bounded rule of \citet{chen2022fully}. All of the rules
trained from a random start are $3.2$ to $4.4$ points below the anchored layer, and a rule with a
hard radius bound diverges. Moving the initial parameters away from the anchor by a random
deviation reduces the accuracy by $0.4$ points at $\sigma = 0.1$ and by $1.1$ points at
$\sigma = 0.2$.

\begin{table}[h]
  \centering
  \caption{Rules for the output radius of the polar fully connected layer on CIFAR-100 with
  ViT-Tiny, one seed. $\sigma$ is the standard deviation of a random initial deviation from the
  anchor.}
  \label{tab:radial}
  \small
  \begin{tabular}{lc}
    \toprule
    Radius rule & Accuracy \\
    \midrule
    Anchored gain, Eq.~\eqref{eq:polar-fc} & $\mathbf{78.45}$ \\
    Anchored gain, $\sigma = 0.1$ & 78.02 \\
    Anchored gain, $\sigma = 0.2$ & 77.36 \\
    Free power of the pre-activation norm & 75.22 \\
    Affine in the input radius, Klein input & 75.06 \\
    Affine in the input radius, unit-direction input & 74.42 \\
    Sigmoid-bounded radius~\citep{chen2022fully} & 74.04 \\
    Hard radius bound & diverged \\
    \bottomrule
  \end{tabular}
\end{table}

\subsection{Expressivity Test}
\label{app:exp-test}
We follow the protocol of \citet[Fig.~2]{vanderklis2026fast}, in which a single fully connected
layer maps a fixed input to a target point at a hyperbolic distance of $1$ to $18$ from the
origin. The layer is trained with SGD at a learning rate of $10^{-3}$ for at most $10^4$
iterations. We report the number of iterations until the geodesic distance to the target falls
below $0.1$, averaged over $3$ seeds. For all Lorentz layers except that of
\citet{chen2022fully}, the loss is the polar distance of Eq.~\eqref{eq:polar-dist}. We run the
layer of \citet{chen2022fully} with its original loss and a learning rate of $0.1$, and the
Poincar\'e layer also uses its original loss.

\begin{figure}[h]
  \centering
  \includegraphics[width=0.75\linewidth]{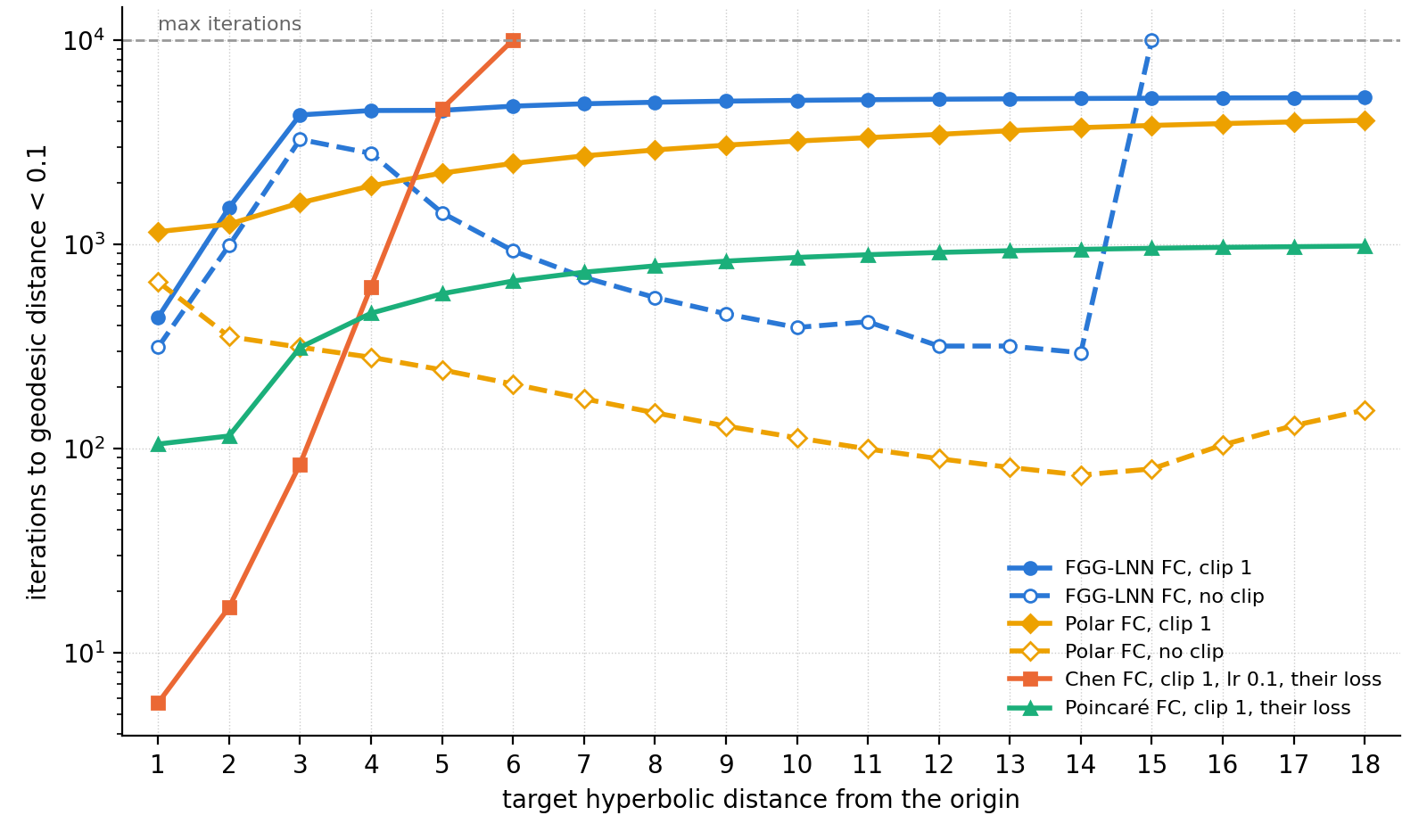}\\
  {\small (a) Float64}\\[4pt]
  \includegraphics[width=0.75\linewidth]{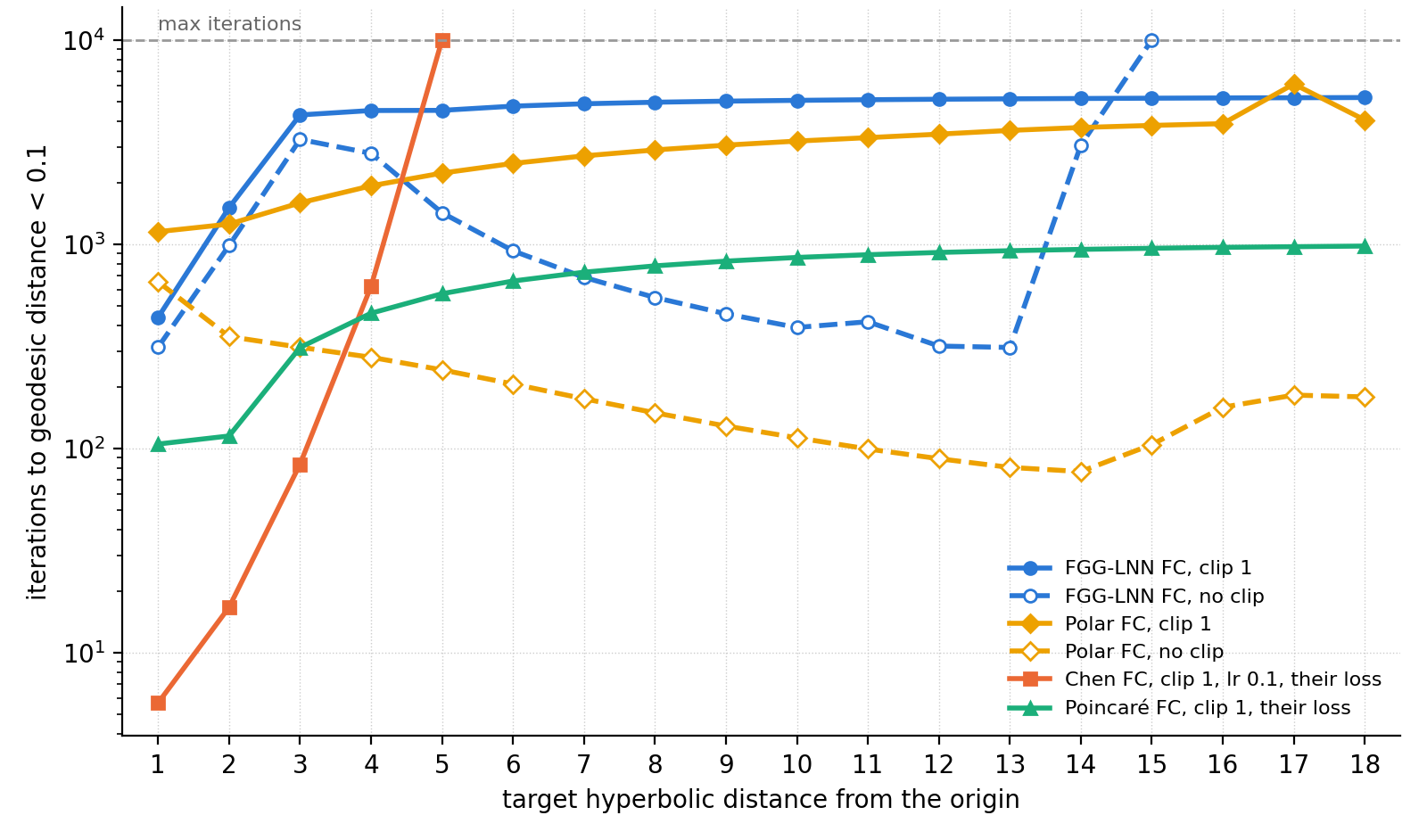}\\
  {\small (b) Float32}
  \caption{Iterations needed by a single fully connected layer to reach a target point, as a
  function of the distance of the target from the origin. We report the mean over $3$ seeds. Runs
  that do not converge are shown at the limit of $10^4$ iterations.}
  \label{fig:exp-test}
\end{figure}

Figure~\ref{fig:exp-test} shows the results. The layer of \citet{chen2022fully} reaches the
iteration limit at radius $6$ in Float64 and at radius $5$ in Float32. The Polar-FC reaches every
target in both precisions. Without clipping, it needs $70$ to $650$ iterations, which is fewer
than any other layer beyond radius $3$. With a clipping threshold of $1$, it needs $1{,}100$ to
$6{,}000$ iterations. FGG-LNN converges at every radius with clipping, but without clipping it
fails from radius $15$ in Float64 and from radius $14$ in Float32.

\subsection{Numerical Precision}
\label{sec:exp-precision}
Table~\ref{tab:precision} reports the single-precision error of the ambient and polar operations
against double precision, measured on clusters of $64$ points with a geodesic spread of $1$. As
predicted by Proposition~\ref{prop:ambient}, the ambient squared distance and centroid deviate from
the true values at $r = 8$, and beyond this radius the ambient centroid returns a point at
radius $9$ for every input. The polar operations stay within a factor of two of the resolution
limit of Proposition~\ref{prop:polar} at every radius.

We also test the low-weight regime of Section~\ref{sec:sensitive}. For one-hot attention rows at
$r \geq 8$, the ambient centroid places the output at radius $9$ or at radius $r + 9.2$ regardless
of the values, and $88$ to $100\%$ of these rows are off by more than $0.5$ in geodesic distance.
The manifold weight decay step of Riemannian AdamW computed with the ambient centroid is off by
$0.1$ at $r = 8$ and increases the parameter radius from $r = 12$ onward. The polar step matches
double precision at every radius.

\begin{table}[h]
  \centering
  \caption{Single-precision error against double precision on clusters of $64$ points with
  geodesic spread $1$ at radius $r$ ($k = 1$). We report the relative error for the squared
  distance and the geodesic error for the weighted centroid and gyro-norm centering. The last
  column is the bound of Proposition~\ref{prop:polar} for $\Delta\theta = 2^{-24}$.}
  \label{tab:precision}
  \small
  \begin{tabular}{rcccccccc}
    \toprule
    & \multicolumn{2}{c}{Squared distance} & \multicolumn{2}{c}{Centroid} & \multicolumn{2}{c}{Gyro-norm centering} & \\
    \cmidrule(lr){2-3} \cmidrule(lr){4-5} \cmidrule(lr){6-7}
    $r$ & Ambient & Polar & Ambient & Polar & Ambient & Polar & Resolution limit \\
    \midrule
     4 & $3 \cdot 10^{-4}$ & $4 \cdot 10^{-7}$ & $2 \cdot 10^{-4}$ & $9 \cdot 10^{-7}$ & $1.2 \cdot 10^{-4}$ & $5.6 \cdot 10^{-6}$ & $5 \cdot 10^{-6}$ \\
     8 & $7 \cdot 10^{-1}$ & $1 \cdot 10^{-5}$ & $9$ & $5 \cdot 10^{-5}$ & $7.0 \cdot 10^{-1}$ & $4.0 \cdot 10^{-4}$ & $2 \cdot 10^{-4}$ \\
    12 & $2 \cdot 10^{3}$ & $7 \cdot 10^{-4}$ & $9$ & $3 \cdot 10^{-3}$ & NaN & $2.0 \cdot 10^{-2}$ & $1 \cdot 10^{-2}$ \\
    \bottomrule
  \end{tabular}
\end{table}

\subsection{Computational Cost}
Table~\ref{tab:cost} reports the time of one forward and backward pass of ViT-Tiny at a batch size
of $128$ on an RTX A5000 under \texttt{torch.compile}. In single precision, the polar model takes
$1.1$ times as long as the ambient Lorentz model and $2.8$ times as long as the Euclidean ViT.
At this input size, the fused attention kernel reduces the time of the attention core from $4.5$ to
$3.0$\,ms. A polar model whose layers pass $(r, \mathbf{u})$ directly agrees with the ambient
model to $10^{-14}$ in double precision.

\begin{table}[h]
  \centering
  \caption{Forward and backward time per training step of ViT-Tiny on CIFAR-100 with batch size
  $128$ on an RTX A5000, compiled. The times include only the model.}
  \label{tab:cost}
  \small
  \begin{tabular}{lc}
    \toprule
    Model & fp32 \\
    \midrule
    Euclidean ViT & 21\,ms \\
    Lorentz ViT, ambient arithmetic & 52\,ms \\
    \method, polar arithmetic & 58\,ms \\
    \bottomrule
  \end{tabular}
\end{table}

\section{Polar Layers in a Convolutional Network}
\label{app:unet}
The Polar-FC of Section~\ref{sec:polar-fc} does not depend on attention. A Lorentz convolution
concatenates the points of each receptive field and applies a fully connected layer to the
result~\citep{bdeir2024hcnn}, so replacing this layer, the batch normalization and the activation
with their polar forms turns a Lorentz CNN into a polar one without changing its structure. We
test this on the medical segmentation benchmark of the Hyperbolic U-Net~\citep{mishra2026hyperbolic}
and compare against the published Euclidean and Poincar\'e U-Nets and a Lorentz U-Net built from
the ambient layers of \citet{bdeir2024hcnn}.

\paragraph{Datasets.}
We use the five datasets and splits of \citet{mishra2026hyperbolic}. These are skin lesions from
ISIC~2016~\citep{gutman2016isic} ($810$ training, $90$ validation and $379$ test images) and
ISIC~2018~\citep{codella2019isic} ($2{,}335$ / $259$ / $1{,}000$), polyps from
Kvasir-SEG~\citep{jha2020kvasir} ($810$ / $90$ / $100$) and from the PraNet
collection~\citep{fan2020pranet} (SANET, $1{,}305$ / $145$ / $798$, where the test set is the union
of CVC-300, CVC-ClinicDB, CVC-ColonDB, ETIS-LaribPolypDB and Kvasir), and breast ultrasound images
from BUSI~\citep{aldhabyani2020busi} ($624$ / $78$ / $78$). Images are zero-padded to a square,
scaled to $[0, 1]$ and resized to $256 \times 256$, without augmentation. We reuse the reference
data loaders, and our training, validation and test image lists are identical to theirs on all
five datasets.

\paragraph{Models.}
All models share the U-Net topology of \citet{mishra2026hyperbolic}, which has four levels with
$8$ channels at the first level and twice as many at each following level, two $3 \times 3$
convolutions per level, max pooling between the encoder levels, and a $1 \times 1$ convolution as
the output layer.
\begin{itemize}
  \item \emph{Euclidean U-Net}~\citep{ronneberger2015unet}: the network of the reference
  implementation, with transposed-convolution upsampling ($486.6$\,K parameters).
  \item \emph{Poincar\'e U-Net}~\citep{mishra2026hyperbolic}: Poincar\'e-ball convolutions with a
  trainable curvature initialized at $-0.1$.
  \item \emph{Lorentz U-Net, ambient}: the convolution, batch normalization and ReLU of
  \citet{bdeir2024hcnn} on points stored in ambient coordinates $(x_0, \mathbf{x}_s)$, with their
  soft radius cap on every convolution input and batch normalization output. Upsampling is
  bilinear, where each new point is the Lorentz centroid of its neighbors. Skip connections are
  joined by Lorentz direct concatenation as in Section~\ref{sec:attention}, and the logits are
  the space part of the logarithmic map of the output layer at the origin ($216.0$\,K parameters).
  \item \emph{Lorentz U-Net, polar}: the same network with each layer replaced by its polar form.
  Each convolution concatenates its receptive field in polar form and applies the Polar-FC of
  Eq.~\eqref{eq:polar-fc}. The batch normalization is that of \citet{bdeir2024hcnn} computed in
  polar form, and the ReLU acts on the space part.
  Pooling, upsampling and concatenation are the same as in the ambient model, and the output layer
  is a polar $1 \times 1$ convolution. The polar model has no radius cap ($215.5$\,K parameters).
\end{itemize}
For the Euclidean and Poincar\'e U-Nets, we take the published numbers of
\citet{mishra2026hyperbolic}.

\paragraph{Training.}
Both Lorentz models use the same recipe. The loss is the sum of the Dice and cross-entropy losses
on a two-class softmax, and we train for $50$ epochs with a batch size of $8$. We use AdamW, with
Riemannian AdamW~\citep{bdeir2025robust} for the manifold parameters, a learning rate of
$10^{-3}$ with cosine decay to $10^{-6}$, and a weight decay of $5 \cdot 10^{-2}$ that is not applied
to scale parameters. On BUSI, which is the smallest
dataset, we use a learning rate of $3 \cdot 10^{-4}$ with $5$ warmup epochs for both models.
Following the reference code, the validation Dice is computed five times per epoch, and the test
score is that of the checkpoint with the best validation Dice.

\paragraph{Metric.}
We report the Dice score as the reference code computes it, which is the mean of the foreground
and background Dice over the two softmax classes. The published numbers in Table~\ref{tab:unet}
use this definition.

\begin{table}[t]
  \centering
  \caption{Medical image segmentation on the benchmark of \citet{mishra2026hyperbolic}. Test Dice
  (mean over foreground and background). The published rows are taken from \citet{mishra2026hyperbolic}, and all values are rounded to
  two decimals as there. The best value of each column is in bold, and ties are both in bold.}
  \label{tab:unet}
  \small
  \setlength{\tabcolsep}{4pt}
  \resizebox{\linewidth}{!}{%
  \begin{tabular}{llcccccc}
    \toprule
    & Model & Params & ISIC 2016 & ISIC 2018 & Kvasir-SEG & SANET & BUSI \\
    \midrule
    \multirow{2}{*}{\emph{published}}
    & Euclidean U-Net & $486.6$\,K & $0.92$ & $\mathbf{0.89}$ & $\mathbf{0.86}$ & $0.78$ & $\mathbf{0.82}$ \\
    & Poincar\'e U-Net & $\approx 487$\,K & $0.91$ & $0.87$ & $0.83$ & $0.76$ & $0.80$ \\
    \midrule
    \multirow{2}{*}{\emph{ours}}
    & Lorentz U-Net, ambient & $216.0$\,K & $0.92$ & $0.81$ & $0.79$ & $0.73$ & $0.71$ \\
    & Lorentz U-Net, polar & $215.5$\,K & $\mathbf{0.93}$ & $\mathbf{0.89}$ & $\mathbf{0.86}$
      & $\mathbf{0.81}$ & $0.78$ \\
    \bottomrule
  \end{tabular}}
\end{table}

\paragraph{Results.}
Table~\ref{tab:unet} shows that the polar U-Net is above the ambient Lorentz U-Net on all five
datasets by $0.01$ to $0.08$ Dice, with the same seeds and $0.5$\,K fewer parameters. It is above
the published Poincar\'e U-Net on four of the five datasets and $0.02$ below it on BUSI. Compared
to the published Euclidean U-Net, it is higher on ISIC~2016 and SANET, equal on ISIC~2018 and
Kvasir-SEG, and $0.04$ lower on BUSI, while using $0.44$ times its parameters.

The polar model is also the more stable of the two Lorentz models. In some cases, training of
the ambient model had to be repeated with gradient clipping and a lower learning rate.

\end{document}